\documentclass[letterpaper, 10 pt, conference]{ieeeconf}  

\IEEEoverridecommandlockouts                              

\title{\LARGE \bf
Automated Plan Refinement for Improving Efficiency of Robotic Layup of Composite Sheets
}

\usepackage{multirow, amsmath, amssymb, graphicx, xcolor, array, url, float, hyperref}
\newcolumntype{C}[1]{>{\centering\arraybackslash}m{#1}} 

\author{Rutvik Patel$^{1}$, Alec Kanyuck$^{1}$, Zachary McNulty$^{1}$, Zeren Yu$^{1}$, \\ Lisa Carlson$^{2}$, Vann Heng$^{2}$, Brice Johnson$^{2}$ and Satyandra K. Gupta$^{1}$
\thanks{$^{1}$Center for Advanced Manufacturing, Viterbi School of Engineering,
        University of Southern California, Los Angeles, CA, USA.
        }%
\thanks{$^{2}$Boeing Research and Technology, Tukwila, WA, USA.
        }%
}

\begin{document}

\maketitle
\thispagestyle{empty}
\pagestyle{empty}
\newcounter{eqrefscnt} 

\newcommand{\eqrefp}[1]{Eq.~(\ref{#1})}
\newcommand{\eqrefs}[1]{%
  \setcounter{eqrefscnt}{0}
  Eqs.~\renewcommand*{\do}[1]{%
    \ifnum\value{eqrefscnt}>0, \fi
    (\ref{##1})
    \stepcounter{eqrefscnt}
  }%
  \docsvlist{#1}
}


\begin{abstract}

The automation of composite sheet layup is essential to meet the increasing demand for composite materials in various industries. However, draping plans for the robotic layup of composite sheets are not robust. A plan that works well under a certain condition does not work well in a different condition. Changes in operating conditions due to either changes in material properties or working environment may lead a draping plan to exhibit suboptimal performance. In this paper, we present a comprehensive framework aimed at refining plans based on the observed execution performance. Our framework prioritizes the minimization of uncompacted regions while simultaneously improving time efficiency. To achieve this, we integrate human expertise with data-driven decision-making to  refine expert-crafted plans for diverse production environments. We conduct experiments to validate the effectiveness of our approach, revealing significant reductions in the number of corrective paths required compared to initial expert-crafted plans. Through a combination of empirical data analysis, action-effectiveness modeling, and search-based refinement, our system achieves superior time efficiency in robotic layup. Experimental results demonstrate the efficacy of our approach in optimizing the layup process, thereby advancing the state-of-the-art in composite manufacturing automation.

\end{abstract}

\section{INTRODUCTION}

\noindent
The growing demand for composite materials in various industrial applications, such as aerospace and manufacturing \cite{compositesmanufacturing2016}, necessitates efficient automation processes for the layup of composite sheets, traditionally done by hand \cite{elkington2015hand}. To meet these rising demands, automation in this area has been introduced in previous works \cite{malhan2019determining, frketic2017automated, schuster2017autonomous, manyar2022visual, malhan2021automated}. Ensuring the quality of the layup process, detecting defects, and implementing corrective measures are essential aspects that have garnered significant attention in the literature \cite{bjornsson2018automated, elkington2017automated, molfino2014design, gambardella2022defects}. Composite sheet layup involves compacting prepreg composite sheets onto a mold by applying force, typically through a tool such as a roller. In robotic layup, the tool is mounted on the end of a robot arm, which manipulates the tool to achieve the desired compaction. When being compacted, composite sheets stretch and deform over the mold. 

A critical aspect of robotic layup lies in the planning phase, where the generation of layup paths is fundamental. Traditional methods base these paths on surface geometry \cite{mcconachie2020manipulating}, but such approaches often fail due to environment variability, leading to uncompacted regions on composite sheets giving rise to defects such as wrinkles, air pockets, bridging, etc \cite{malhan2021automated}. Draping plans that are generated to work well under certain conditions may not work well in different conditions because of the following two factors. First, once the material is removed from the refrigerator its properties change over time. Second, the ambient conditions may differ from one location to another. Therefore, a draping plan that works well in a particular situation may not work well in a different situation. For example, it may require execution of several additional paths to remove uncompacted regions caused by the stiffening of the material and therefore may increase the cycle time. Intervention controllers have been proposed to address uncompacted regions but they too lead to increased cycle times \cite{manyar2021simulation}. Simulation-based planning approaches are impractical due to the need for frequent recalibration to environmental conditions or updating the material models. Similarly, relying solely on human experts to modify the plans is not feasible is many situations. Our approach refines human expert plans through experimentation without requiring expert intervention. We need an approach that can automatically refine draping plans based on their observed performances.

\begin{figure}[!t]
    \centering
    \includegraphics[width=0.48\textwidth]{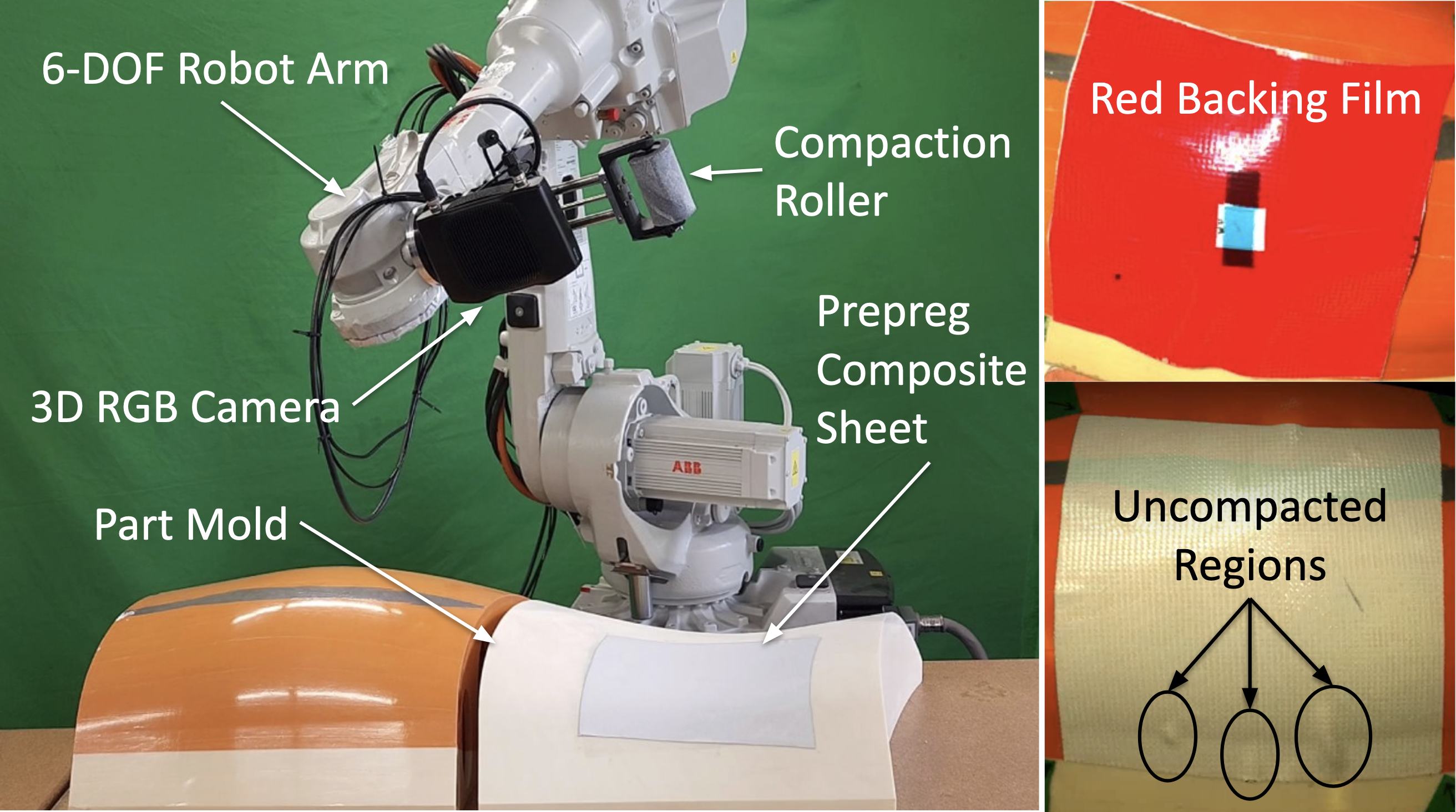}
    \caption{Operational setup for composite sheet layup, comprising a robot arm, 3D RGB camera, part mold, composite sheet, and compaction roller. The composite sheet features red backing film on both sides, which is removed during layup. Uncompacted regions may develop during the layup, requiring correction.}
    \label{fig:operational_setup}
\end{figure}

Our experiments reveal a consistent need for a large number of corrective paths during the correction cycles, nearly equal to the number of initial paths, emphasizing the need for re-evaluating the initial path generation. In response to these challenges, we focus on refining the initial layup paths in the expert-crafted plans to minimize uncompacted regions. Our framework presents a comprehensive approach to refining draping plans, focusing on time efficiency and eliminating uncompacted regions. By combining human decision-making in experiment design with the system's learning capability, we minimize the number of layup paths required, enhancing efficiency and accuracy in robotic layup of composite sheets under varying conditions achieving superior results compared to purely expert-crafted plans in terms of time efficiency and accuracy. 

The novel contributions of this paper are twofold:
\begin{itemize}
    \item First, we have developed a learning methodology that allows us to assess the effectiveness of different types of draping actions in the formation and removal of uncompacted regions from a limited number of experimental data.
    \item Second, we have developed a search process that allows us to compute refined plans from the models built from the experimental data.
\end{itemize}

\section{RELATED WORKS}
\noindent
\textbf{Trajectory planning for sheet layup:} Various trajectory planning approaches for composite sheet layup have been explored in the literature \cite{malhan2021automated}. \cite{ehinger2014robot} involves offline path planning, where experts define layup curves used by planners to generate roller trajectories on the part. However, this method is inefficient as it relies on expert input for precise curve definition on the mold. Another approach in \cite{zhang2018optimizing} involves the use of deflection curves to define surface paths on a base geodesic path. \cite{yan2014accurate} discusses a surface-plane intersection approach to establish an initial path on the part surface, with subsequent paths generated as offsets from this initial path. Our geometry-based planner utilizes this strategy to dynamically generate new paths as needed. Additionally, algorithms have been proposed for generating initial paths and subsequent offsetting trajectories on part surfaces represented by point cloud inputs, accommodating both open and closed surfaces in \cite{wang2023general}. \cite{gao2018optimal} discusses a discretization approach coupled with directed graph search to address trajectory planning. Prior works such as \cite{manyar2021simulation, chen2023multisensor} propose corrective actions aimed at rectifying defects during the layup process. Other works have investigated various strategies for defect detection and correction. These include tactile sensing models \cite{elkington2021real} and vision-based techniques \cite{manyar2022synthetic, tang2022process, chen2021rapid}. Corrective actions aimed at rectifying defects during layup have been proposed in prior works such as \cite{manyar2021simulation, chen2023multisensor}. In this work, we build upon the groundwork laid by vision-based inspection and correction techniques, integrating them within the robotic layup process \cite{manyar2021simulation}.
\\
\textbf{Planning for Manipulating Deformable Objects:} Recent studies have covered a range of research directions to address the problems in deformable object manipulation (DOM) \cite{zhu2022challenges}. \cite{yan2021learning} focuses on optimizing visual representation and dynamics models, \cite{mcconachie2020manipulating} proposes a framework integrating geometry-based planning and local control for deformable object manipulation without high-fidelity simulation, and \cite{mcconachie2020bandit} uses multiple models due to the lack of high fidelity simulations to pick and choose from appropriate models based on their utility for a given task. However, the inherent uncertainties and variations in composite materials pose difficulties for visual representation optimization, especially with limited experimental data and may limit the effectiveness of geometry-based planning methods. In consideration of these challenges, we conclude that it is necessary to build a process model that allows for decision making to refine the initial set of plans.
\\
\textbf{Process model building:} \cite{yan2021learning} employs contrastive learning to enhance predictive modeling of deformable objects for improved plannable representations but relies on gathering random data, limiting practical application in our case. Gaussian Process Regression (GPR) in \cite{hu2018three} capture material shape variability during manipulation. \cite{caccamo2016active} explores online GPR and Position-based Dynamics (PBD) for autonomous control of object position and shape, constructing probabilistic models facilitating real-time estimation of deformations and control command generation. Learning-based feedback controllers in \cite{thach2022learning, jia2018learning} optimize servo control for diverse tasks based on observed visual features. Additionally, \cite{liu2023deformer} introduces novel neural network architecture enabling precise shape manipulation from partial-view point clouds, even for objects with novel material properties.

\section{PROBLEM FORMULATION}

\subsection{Background}
\noindent
The layup process involves a composite prepreg sheet sandwiched between two backing films, both of which will eventually be removed, to be compacted onto a mold surface using a roller. This is also referred to as draping. It follows a pre-defined draping plan that outlines the sequence of actions to be taken. Prior to executing the draping plan, the backing film is peeled off from the side designated for compaction against the surface. Subsequently, a robotic arm-mounted roller is used to compact the sheet onto the mold surface. The backing film on the top surface is peeled off as necessary during the layup. Upon completion of the layup, a vision-based correction controller is used to inspect the sheet and fix the faults. Sections that have not adhered adequately to the surface, appearing as bubbles or wrinkles, are identified through point cloud captures of the surface mold. These uncompacted regions are then systematically removed by moving the roller on top of them to ensure that the final quality of the layup meets the standards.

\subsection{Robotic Cell for Composite Sheet Layup}
\noindent
Fig. \ref{fig:operational_setup} illustrates the various components involved in our cell setup for robotic layup of composite sheets. The layup process is executed on a 6 DOF ABB IRB 2600 industrial robot arm. A specialized 3D printed tool, is affixed to the robot arm. This tool incorporates a dual guide rod pneumatic cylinder, with a foam roller mounted on the end for compaction. The stiffness of the roller can be adjusted by modulating the air pressure supplied to the cylinder. Additionally, mounted to this 3D printed tool is a Zivid 2 3D camera, capable of capturing precise point cloud data.

\subsection{Terminology}
\noindent
A draping plan of length $m$ is an ordered set of actions, denoted by $\mathcal{D} = \langle (\text{type}, \text{arguments})_i \mid i \in \{t_1, t_2, \ldots, t_m\} \rangle$, designed to guide the layup process. Let $\tau$ represent the set of waypoints on the surface: $\{ p_0, \ldots \}$. We define five types of actions that can be incorporated in a draping plan:

\begin{itemize}
	\item \textbf{Path Action:} The path action is modeled by the function $\mathcal{P}: \{1, \ldots, n\} \rightarrow \tau$. It is denoted as $(\text{path}, i)$ in a plan, where $i \in \{1, \ldots, n\}$ represents the selection of a single path from the set of $n$ radially outward paths on the sheet constructed using a purely geometry based planner.
	\item \textbf{Peel Action:} The peel action is modeled by the function $\mathcal{O}: \phi \rightarrow \tau_{\text{peel}}$. It is Denoted as $(\text{peel}, \phi)$ in a plan, indicating the peeling off of the backing film of the sheet.
	\item \textbf{Capture Action:} The capture action is modeled by the function $\mathcal{C}: \phi \rightarrow p_{\text{capture}}$. It is Denoted as $(\text{capture}, \phi)$ in a plan, indicating a camera capture during the layup process.
	\item \textbf{End Action:} The end action is modeled by the function $\mathcal{E}: \phi \rightarrow \phi$. It is denoted as $(\text{end}, \phi)$ in a plan, indicating that the control is surrendered completely to the correction controller.
	\item \textbf{Refinement Action:} The refinement action is modeled by the function $\mathcal{T}: n \in \mathbb{I}^+ \rightarrow \{\tau_1, \tau_2, \ldots, \tau_n\}$. It is denoted as $(\text{refinement}, n)$ in a plan, where $n \in \mathbb{I}^+$ represents the number of refined paths to be generated and inserted into the plan at this stage.
\end{itemize}

The action space $\mathbb{A}$ is thus the set $\{\mathcal{P} \times \{1, \ldots, n\}\} \cup \{\mathcal{O} \times \phi\} \cup \{\mathcal{C} \times \phi\} \cup \{\mathcal{E} \times \phi\} \cup \{\mathcal{T} \times \mathbb{I}^+\}$. Additionally, $\mathcal{D}$ adheres to constraints defined by $\mathcal{V}$. In our framework, constraints are divided into two categories: relative constraints $\mathcal{V}_{rel}$ and absolute constraints $\mathcal{V}_{abs}$. Refer to table \ref{table:constraints} for the description.

\begin{table}[!ht]
\centering
\begin{tabular}{|m{8cm}|}
\hline
\textbf{Constraint Type: Relative Constraints ($V\textsubscript{rel}$)} \\
\hline
\textbf{Notation}: ($\alpha$, $\beta$, $\gamma$, $\lambda$) \\
\textbf{Description}: \\
\begin{itemize}
    \item \textbf{$\alpha$}: Action under constraint
    \item \textbf{$\beta$}: Reference action
    \item \textbf{$\gamma$}: Relationship between $\alpha$ and $\beta$
    \begin{itemize}
        \item \textbf{$>$}: $\alpha$ occurs more than $\lambda$ actions after $\beta$
        \item \textbf{$=$}: $\alpha$ occurs exactly $\lambda$ actions after $\beta$
        \item \textbf{$<$}: $\alpha$ occurs within $\lambda$ actions after $\beta$
    \end{itemize}
    \item \textbf{$\lambda$}: Number of intermediary actions
\end{itemize}
\\ \hline
\textbf{Constraint Type: Absolute Constraints ($V\textsubscript{abs}$)} \\
\hline
\textbf{Notation}: ($\alpha$, $\gamma$, $\lambda$) \\
\textbf{Description}: \\
\begin{itemize}
    \item \textbf{$\alpha$}: Action under constraint
    \item \textbf{$\gamma$}: Relationship between $\alpha$ and $\lambda$
    \begin{itemize}
        \item \textbf{$>$}: $\alpha$ occurs more than $\lambda$ times in the plan
        \item \textbf{$=$}: $\alpha$ occurs exactly $\lambda$ times in the plan
        \item \textbf{$<$}: $\alpha$ occurs less than $\lambda$ times in the plan
    \end{itemize}
    \item \textbf{$\lambda$}: Frequency count
\end{itemize}
\\ \hline
\end{tabular}
\caption{Summary of Constraints}
\label{table:constraints}
\end{table}

\subsection{Problem Statement}
\noindent
The central challenge addressed in this study involves enhancing the layup process efficiency through the creation of a refined draping plan. Utilizing a set of empirical data generated from executing a range of expert-crafted draping plans, the aim is to refine these initial plans, surpassing their efficacy in terms of both time efficiency and accuracy.

\begin{figure}[htbp]
    \centering
    \includegraphics[width=0.45\textwidth]{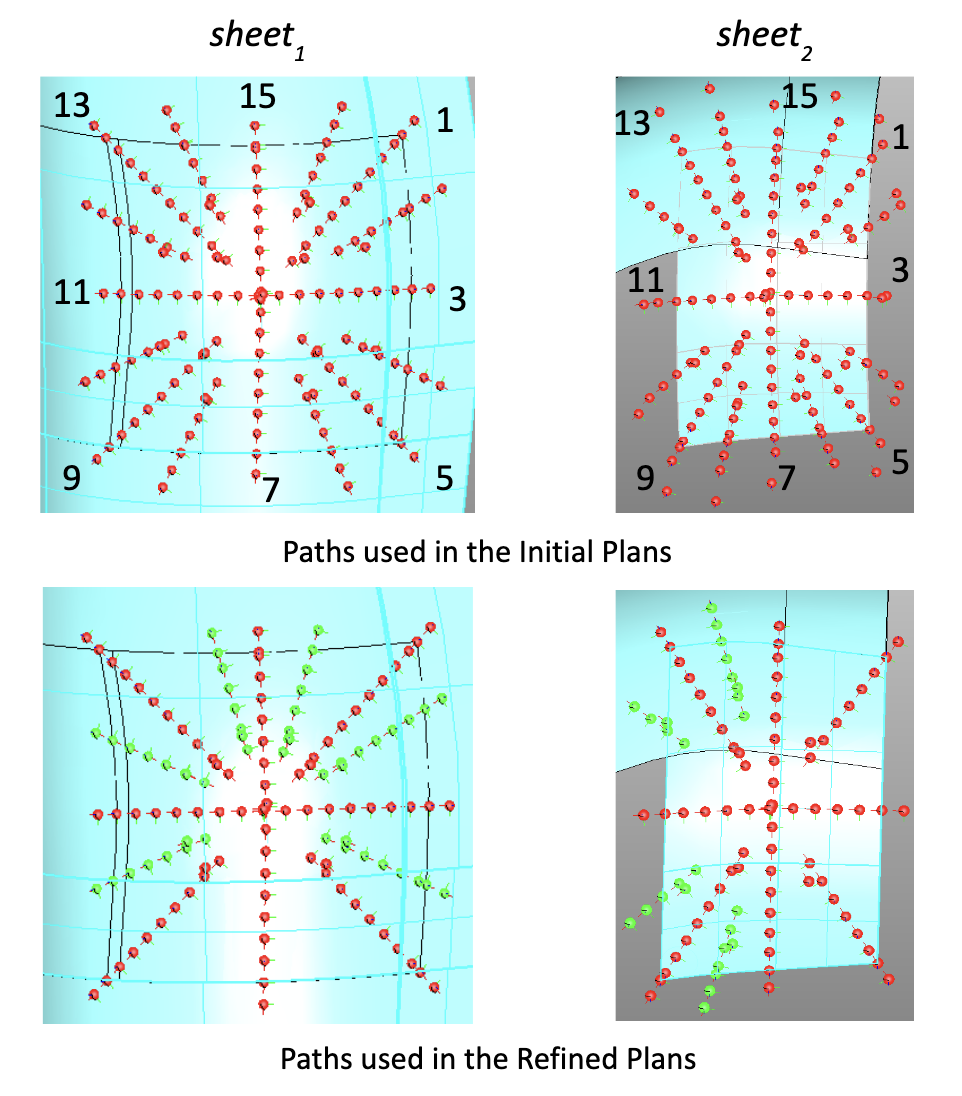}
    \caption{In the initial plans, paths are generated by the geometry-based planner. The red paths are numbered clockwise from 1 to 16, starting from the top-right corner of the sheets. In the refined plans, paths comprises of those from the geometry-based planner (red) and additional paths generated using the refinement action (green). The selection and ordering of paths varies between draping plans.}
    \label{fig:paths_used}
\end{figure}

\begin{figure*}[!h]
    \centering
    \includegraphics[width=1.0\textwidth]{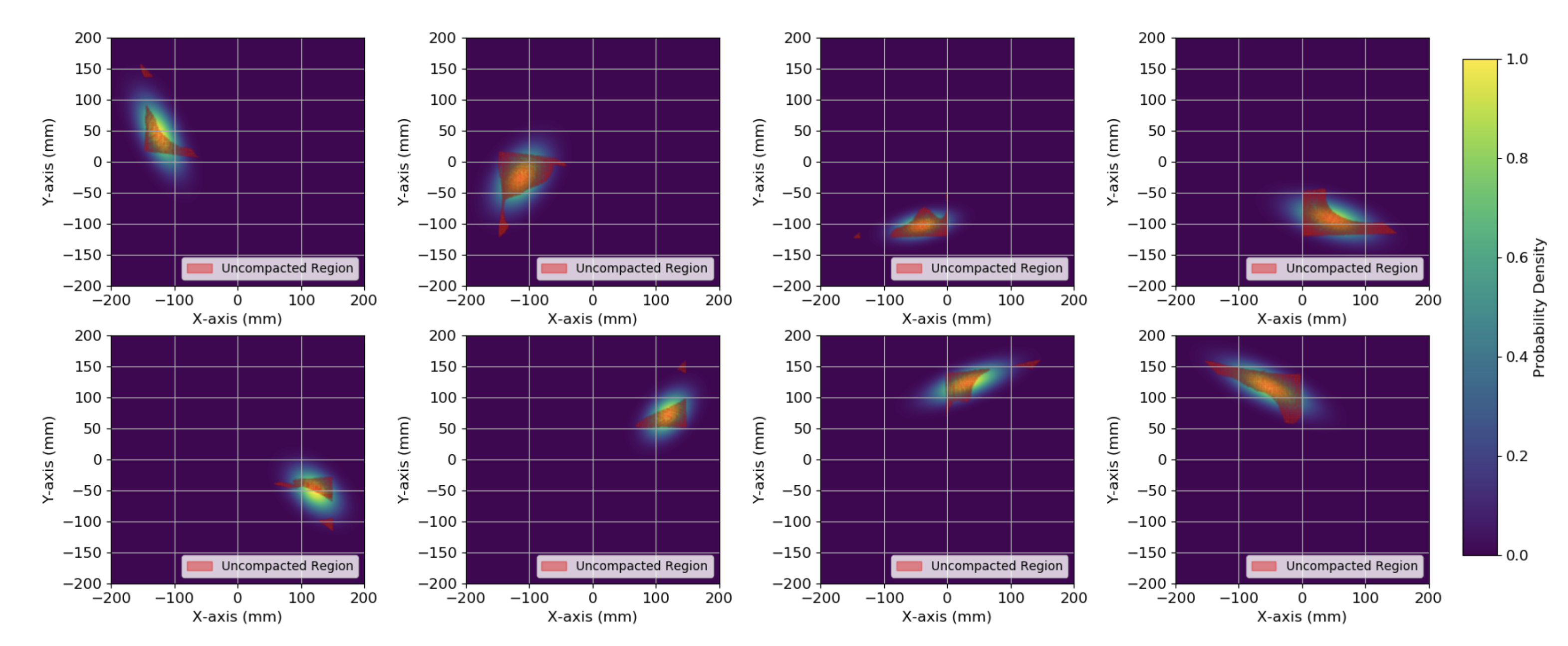}
    \caption{2D visualization of $G_{1i}$ across all eight sheet sectors. The actual uncompacted region is highlighted in red, with the color scale indicating the Gaussian probability mass corresponding to the centroid of the uncompacted region.}
    \label{fig:sector_centroid}
\end{figure*}

The decision-making process operates over the action space $\mathbb{A}$, which includes paths derived from experimental data through the refinement action, paths derived from geometry-based planner through the path action, and specific actions such as peel, capture, and end. For the path action, we have $n=16$. Fig. \ref{fig:paths_used} shows these paths. The objective is to select and order the actions into a draping plan $\mathcal{\hat{D}}$ to minimize efforts based on a cost function defined over our internal state representation of the sheet, subject to constraints defined by $\mathcal{V}$. We discuss the details in the following sections.
\\
\textbf{Assumptions}: We make the following assumptions for this study:
\begin{itemize}
    \item We assume that the layup process is Markovian – the current state of the sheet alone is sufficient to make a decision for future actions.
    \item We assume that the layup process is typically unaffected by major external disruptions, except for the inherent randomness in the operational setting. This randomness may be caused by variations in sheet material properties, environmental factors such as temperature, pressure, and humidity, as well as minor fluctuations in the behavior of the robotic arm, including slight alterations in speed, angle, and impact points.
\end{itemize}

\subsection{Overview of Approach}
\noindent
This section provides a comprehensive overview of our approach. Our approach is divided into sequential phases which are discussed below.

\begin{enumerate}
    \item \textbf{Generate Plans Based on Expert's Suggestions:} Our approach begins with the human experts drafting initial draping plans. Drawing upon their domain knowledge and experience, the experts draft alternative initial draping plans. The experts consider surface geometry, sheet material type, sheet behavior, environmental factors, etc. to assemble a diverse set of plans covering a range of approaches. This human-driven input serves as the foundation for subsequent analysis and improvements toward an improved plan.

    \item \textbf{Execute Plans and Process Data:} In this phase, the generated plans are executed with the robot in the operational environment to perform the layup operation. An experiment is designed for every draping plan and during execution data such as point cloud capture is collected after every path executed by robot along with the state of the sheet at the end of the plan. The point clouds are processed to undergo filtering to identify uncompacted regions, followed by discretization and transformation into our internal state representation. Please refer to section~\ref{sec:action} for more details. This data is used in the next phase to characterize sector-wise distributions of changes for every action. 

    \item \textbf{Characterize Action-Effectiveness:} In this phase we construct a model to characterize action-effectiveness. By analyzing the cause-and-effect relationships within the executed plans through the before and after states of the sheet with respect to an action, we learn the impact of individual actions on various sections of the sheet. This modeling step is essential in quantifying the effectiveness of various draping plans and the individual actions in them, and in refining the understanding of successful planning components for both experts and our algorithm. Please refer to section~\ref{sec:action} for more details.
    
    \item \textbf{Perform Search to Refine Draping Plans:} We use search as a technique to construct a draping plan. This phase involves exploring the action space using the action-effectiveness model generated in the previous step. The search problem is modeled as a search in a tree with nodes being actions connected by branches with cost determined by the effectiveness score of the action at that stage and the penalty to execute the action. The search terminates when we converge at a stage where no action can improve the state of the sheet or at the end of the horizon for the plan. The path followed in the tree represents the refined draping plan. See Section~\ref{sec:search} for more details.

    \item \textbf{Execute and Evaluate Refined Plan:} With the optimal plan in hand, we design an experiment and perform it while capturing the same data as done for the initial plans. Next, we assess the plan's performance against initial plans and quantify the improvements in terms of the reduction in the average number of paths required in the layup including the paths generated by the correction controller.
\end{enumerate}

\section{Learning Action Effectiveness} 
\label{sec:action}
\noindent
\textbf{State Representation}: The sheet is discretized into sectors $S_i$ for $i \in \{1, \ldots, k\}$. These sectors encircle the center of the sheet. We use $k=8$. Corresponding to each sector are two multivariate Gaussian distributions $G_{1i}$ and $G_{2i}$. $G_{1i} = \mathcal{N}(\mu_{1i}, \Sigma_{1i})$ represents the spatial distribution of uncompacted regions in the sector where \(\mu_i \in \mathbb{R}^3\) represents \(x\), \(y\) with respect to the sheet center, and height of the uncompacted region, and \(\Sigma_i\) is the corresponding $3 \times 3$ covariance matrix. Conditioned on \(G_{1i}\), \(G_{2i} = \mathcal{N}(\mu_{2i}, \Sigma_{2i})\) represents the dimensions and orientations of the uncompacted regions in the sector. The uncompacted regions of each sector are fitted into ellipses, capturing the size along the major and minor axes and the orientation of the ellipses. Here, \(\mu_{2i} \in \mathbb{R}^3\) represents the lengths of the major and minor axes of the ellipse and its orientation, and \(\Sigma_{2i}\) the corresponding $3 \times 3$ covariance matrix. Fig.  \ref{fig:sector_centroid} shows \(x\) and \(y\) components of \(G_{1i}\) for different sectors from one of our experiments.
\\
\textbf{Action Effectiveness Modelling}: We perform experiments with different draping plans to study their effects on the sheet. We model the changes caused by an action by computing the changes caused by it on the distributions $G_{1i}$ and $G_{2i}$ in each sector. The objective is to learn the per sector changes caused by an action using the before distributions $G_{1i}^{t_{j-1}}$ and $G_{2i}^{t_{j-1}}$, and after distributions $G_{1i}^{t_j}$ and $G_{2i}^{t_j}$ for each action-sector pairing \((a_j, S_i): a_j \in \mathcal{D}, 1 \leq i \leq k\). The experimental data is used to create distributions of the changes observed in each sector per action. \eqrefs{eq:1,eq:2,eq:3} are used to model the changes for every sector over multiple experiments of different draping plans.

\begin{gather}
    \delta^{t_j}_{S_i} = 
    \begin{bmatrix}
        \delta^{t_j}_{x, S_i} \\
        \delta^{t_j}_{y, S_i} \\
        \delta^{t_j}_{h, S_i} \\
        \delta^{t_j}_{a, S_i} \\
        \delta^{t_j}_{b, S_i} \\
        \delta^{t_j}_{\theta, S_i}
    \end{bmatrix} = 
    \begin{bmatrix}
        \mu_{x, 1i}^{t_j} - \mu_{x, 1i}^{t_{j - 1}} \\
        \mu_{y, 1i}^{t_j} - \mu_{y, 1i}^{t_{j - 1}} \\
        \mu_{h, 1i}^{t_j} - \mu_{h, 1i}^{t_{j - 1}} \\
        \mu_{a, 2i}^{t_j} - \mu_{a, 2i}^{t_{j - 1}} \\
        \mu_{b, 2i}^{t_j} - \mu_{b, 2i}^{t_{j - 1}} \\
        \mu_{\theta, 2i}^{t_j} - \mu_{\theta, 2i}^{t_{j - 1}} \\[10pt]
    \end{bmatrix} \label{eq:1} \\
    [\mathcal{U}^{t_j}_{1, S_i}]_{ll} = \begin{cases}
    \mathcal{H}([\Sigma_{1i}^{t}]_{ll} - [\Sigma_{1i}^{t_{j - 1}}]_{ll}), & \text{if } l \in \{1, 2, 3\} \\
    0, & \text{otherwise}
    \end{cases} \label{eq:2} \\
    [\mathcal{U}^{t_j}_{2, S_i}]_{ll} = \begin{cases}
    \mathcal{H}([\Sigma_{2i}^{t}]_{ll} - [\Sigma_{2i}^{t_{j - 1}}]_{ll}), & \text{if } l \in \{1, 2, 3\} \\
    0, & \text{otherwise}
    \end{cases} \label{eq:3} \\
    \nonumber
    \mathcal{H}(x) = 
    \begin{cases}
        1 & \text{if } x > 0 \\
        -1 & \text{if } x \leq 0
    \end{cases}
\end{gather}

\begin{figure}[htbp]
    \centering
    \includegraphics[width=0.45\textwidth]{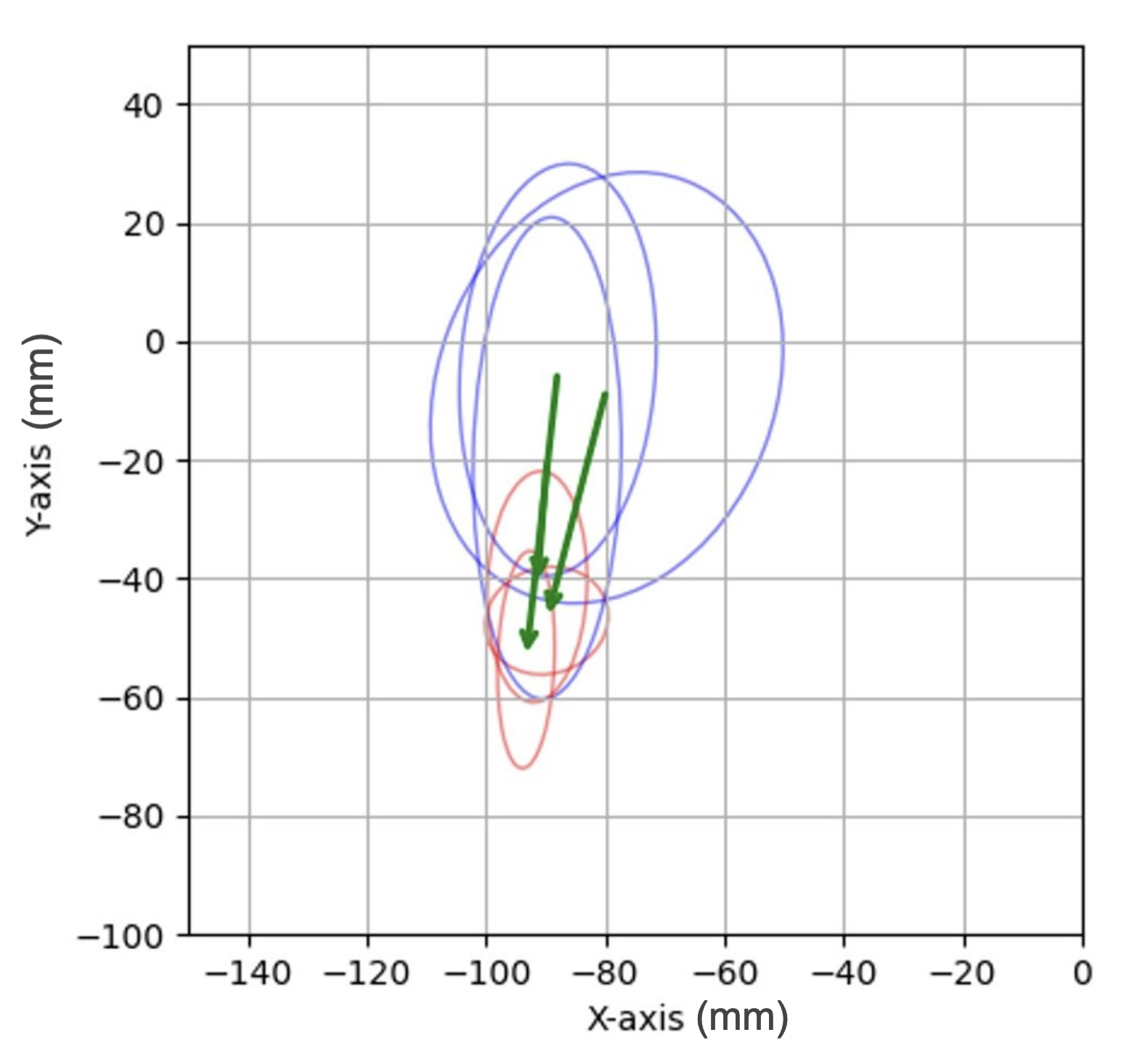}
    \caption{Visualization of changes within a single sector resulting from a $path$ action. Ellipses represent uncompacted regions. Blue indicates the original uncompacted region, while red denotes the updated size, location, and orientation of this region after the action. The centroid shift is depicted by the green arrow. The change in spread is illustrated by ellipses, which are fitted to the $2\sigma$ values of the major and minor axes and oriented at the mean angle $\theta$ from $G_{2i}$ before and after the action.}
    \label{fig:action_changes}
\end{figure}

Specifically, \(G_{1i}\) is used for learning how a given action impacts the position of the center within a sector. This allows the system to model how an action displaces the centroid of the uncompacted region within the sector, using differences \(\delta_x\) and \(\delta_y\). Additionally, \(G_{1i}\) informs the system about variations in height observed after an action to model the effectiveness of that action in compacting the uncompacted region on the mold surface, using \(\delta_h\). Conversely, \(G_{2i}\) is employed to learn alterations in the spread and orientation of the uncompacted region within a sector following a specific action. The system models how the action influences the spatial distribution and orientation of the uncompacted region using \(\delta_a\), \(\delta_b\), and \(\delta_\theta\). This enables the system to learn the evolution of the shape of the uncompacted regions within each of the sectors which is useful for planning. Fig. \ref{fig:action_changes} shows visually the changes in \(G_{1i}\) and \(G_{2i}\) for one of the sectors. Additionally, we keep track of $3 \times 3$ diagonal matrices $\mathcal{U}_1$ and $\mathcal{U}_2$ derived from the covariance matrices of the Gaussians to indicate if the change we observed is a positive change that reduces uncertainty or not. We obtain the combined distribution over all draping plans in \eqrefp{eq:5}.

\begin{gather}
    \Delta = \bigcup_{\{\mathcal{D}_1, \mathcal{D}_2, \ldots\}} \bigcup_{t \in \{t_1, t_2, \ldots, t_m\}} \langle \delta^t_{S_i} : i \in \{1, \ldots, k\} \rangle \label{eq:5}
\end{gather}

\begin{table*}[!t]
\centering
\caption{Performance of Expert-Crafted Initial Plans and Refined Plans}
\label{tab:results}
\begin{tabular}{|C{1.2cm}|C{1.6cm}|C{1.6cm}|C{1.6cm}|C{2.4cm}|C{2cm}|C{1.4cm}|C{1.8cm}|} 
\hline
\textbf{Sheet}  & \textbf{Plan} & \textbf{Paths in Plan} & \textbf{Trial Number} & \textbf{Number of Correction Cycles} & \textbf{Correction Paths} & \textbf{Total Paths} & \textbf{Average Paths} \\ \hline
\multirow{9}{*}{\(sheet_1\)} & \multirow{3}{*}{$\mathcal{D}_{1}$} & \multirow{3}{*}{16} & Trial 1 & 5 & 17 & 33 & \multirow{3}{*}{37.0} \\ \cline{4-7}
& & & Trial 2 & 7 & 30 & 46 & \\ \cline{4-7}
& & & Trial 3 & 5 & 16 & 32 & \\ \cline{2-8}
& \multirow{3}{*}{$\mathcal{D}_{2}$} & \multirow{3}{*}{16} & Trial 1 & 7 & 29 & 45 & \multirow{3}{*}{34.3} \\ \cline{4-7}
& & & Trial 2 & 2 & 12 & 28 & \\ \cline{4-7}
& & & Trial 3 & 4 & 14 & 30 & \\ \cline{2-8}
& \multirow{3}{*}{$\mathcal{\hat{D}}$} & \multirow{3}{*}{14} & Trial 1 & 2 & 5 & 19 & \multirow{3}{*}{20.0} \\ \cline{4-7}
& & & Trial 2 & 2 & 5 & 19 & \\ \cline{4-7}
& & & Trial 3 & 3 & 8 & 22 & \\ \hline
\multirow{9}{*}{\(sheet_2\)} & \multirow{3}{*}{$\mathcal{D}_{1}$} & \multirow{3}{*}{16} & Trial 1 & 2 & 8 & 24 & \multirow{3}{*}{25.3} \\ \cline{4-7}
& & & Trial 2 & 2 & 9 & 25 & \\ \cline{4-7}
& & & Trial 3 & 3 & 11 & 27 & \\ \cline{2-8} 
& \multirow{3}{*}{$\mathcal{D}_{2}$} & \multirow{3}{*}{16} & Trial 1 & 2 & 12 & 28 & \multirow{3}{*}{27.3} \\ \cline{4-7}
& & & Trial 2 & 3 & 9 & 25 & \\ \cline{4-7}
& & & Trial 3 & 3 & 13 & 29 & \\ \cline{2-8}
& \multirow{3}{*}{$\mathcal{\hat{D}}$} & \multirow{3}{*}{12} & Trial 1 & 1 & 5 & 17 & \multirow{3}{*}{16.3} \\ \cline{4-7}
& & & Trial 2 & 3 & 5 & 17 & \\ \cline{4-7}
& & & Trial 3 & 2 & 3 & 15 & \\ \hline
\end{tabular}
\end{table*}

\section{Generating Refined Plans}
\label{sec:search} 

\noindent
Given an initial state and the action effectiveness distribution, we use tree-search as a technique to find a refined sequence of actions that minimizes total paths. Actions along with stochastic moves generated by nature, when executed, modify the internal state. The primary objective is to identify \(\hat{D}\), a refined sequence of actions minimizing a cost function. The problem is formulated, based on the sheet's internal state in \eqrefp{eq:cost}.

\begin{align}
\mathcal{C}(\mathcal{\hat{D}}) &= \sum_{i=1}^{n} (c(a_i) + f(\mathcal{X}_i)) \label{eq:cost}\\
\nonumber \mathcal{\hat{D}} &= \arg\min_{\mathcal{D} \in A^m} \mathcal{C}(\mathcal{D}) \\
\nonumber \text{subject to:} \\
\nonumber &\text{Relative Constraints: } \mathcal{V}_{rel} \\
\nonumber &\text{Absolute Constraints: } \mathcal{V}_{abs}
\end{align}

where, \(\mathcal{X}_i\) is the state at the end of the \(i\)-th action \(a_i\) in the draping plan, \(c(a_i)\) represents the cost associated with taking action on state \(S_i\) and \(f(\mathcal{X}_i)\) represents the utility associated with the sheet state \(\mathcal{X}_i\).

The process starts by initializing the internal sheet representation using the point cloud capture of the sheet at the beginning of the layup from historical data. We sample actions from $\mathbb{A}$ which are used to propagate the internal state along with process noise. Represented by branches in the tree, each action adjusts the internal state of the sheet using $\Delta$ for each action-sector pair. At each stage, we enforce the constraints on \(\mathbb{A}\) before sampling actions. We use the following constraints: $V_{rel}=\{(\mathcal{E}, \mathcal{P}, >, 0), (\mathcal{E}, \mathcal{O}, >, 0), (\mathcal{E}, \mathcal{C}, >, 0), (\mathcal{E}, \mathcal{T}, >, 0)\}$ and $V_{abs}=\{(\mathcal{O},>,0), (\mathcal{E},>,0), (\mathcal{T},=,1), (\mathcal{C},=,1)\}$.

The search continues until either the maximum allowed length is reached or the internal state converges, signifying that no action can further improve the sheet state. Throughout the search, heuristics, guided by the state representation and $\mathcal{U}_1$ and $\mathcal{U}_2$, assist the system in selecting the best \(b_f\) (branching factor) actions for further exploration. The choice of actions is based on estimates of \(\Sigma_{1i}\) and \(\Sigma_{2i}\), along with the values of $a$, $b$, and $h$ after propagating the state with the actions. For each of the best $b_f$ actions, sub-trees are explored up to a specified depth \(d_f\) (depth factor), to estimate the expected value of executing a particular action. The action leading to the sub-tree with the best expected state at depth \(d_f\) is then included in the draping plan. For every refinement action in the plan, we use the parameters \(x\), \(y\), \(a\), \(b\) and \(\theta\) to generate a path. Values of \(x\), \(y\), \(a\) and \(b\) are used for the starting and ending positions of the path while, \(a\), \(b\) and \(\theta\) inform the direction of the path on the surface. To manage computational complexity, the hyper-parameters \(b_f\) and \(d_f\) can be adjusted. 

\section{RESULTS}
\begin{figure}[htbp]
    \centering
    \includegraphics[width=0.45\textwidth]{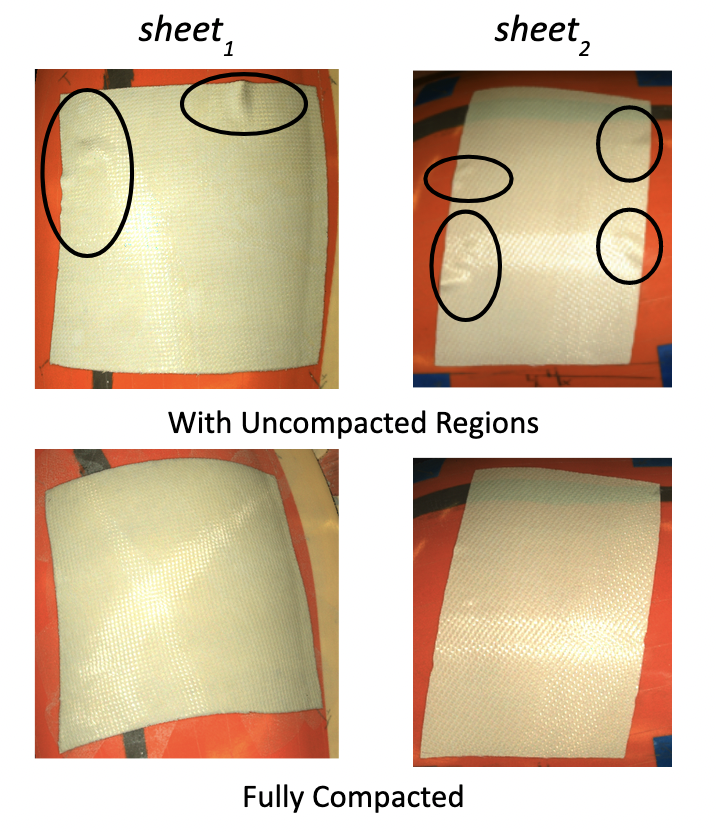} 
    \caption{Comparison between $sheet_1$ and $sheet_2$ showing their uncompacted regions before the completion of the layup process and after being fully compacted at the end of the layup process.}
    \label{fig:sheet_status}
    \vspace{-0.2in}
\end{figure}

\noindent
\textbf{Experiments and Data Collection:} For our experimentation, we use a 3D printed mold designed to mimic the topology of an aircraft part with significant curvature. As shown in Fig. \ref{fig:sheet_status}, we use two different types of sheets — a squarish sheet \(sheet_1\) and a rectangular sheet \(sheet_2\) — both made of a tightly adherent, solvent-based material. These sheets are laid up on the mold in different positions and orientations. For both types of sheets, experts provide us with two initial draping plans as specified in table \ref{tab:initial-plan}. The paths in these plans are depicted in Fig. \ref{fig:paths_used}. Each combination of sheet and draping plan undergoes three experiments. Throughout these experiments, point cloud data is captured after every action unlike the typical layup procedures where capture occurs only for the capture action $\mathcal{C}$. In total, 12 experiments are conducted using the initial plans. Table \ref{tab:results} summarizes the findings from these experiments.
\vspace{-2mm}

\begin{table}[h]
\centering
\caption{Initial Draping Plans for both Sheets}
\label{tab:initial-plan}
\begin{tabular}{|c|c|c|}
\hline
\textbf{Time Step (t)} & \textbf{$\mathbf{D_1}$} & \textbf{$\mathbf{D_2}$} \\
\hline
$t_{1}$ & (path, 15) & (path, 3) \\
$t_{2}$ & (path, 9) & (path, 11) \\
$t_{3}$ & (path, 5) & (path, 7) \\
$t_{4}$ & (path, 13) & (path, 15) \\
$t_{5}$ & (path, 3) & (path, 1) \\
$t_{6}$ & (path, 7) & (path, 9) \\
$t_{7}$ & (path, 11) & (path, 5) \\
$t_{8}$ & (path, 1) & (path, 13) \\
$t_{9}$ & (path, 2) & (path, 2) \\
$t_{10}$ & (path, 4) & (path, 4) \\
$t_{11}$ & (path, 6) & (path, 6) \\
$t_{12}$ & (path, 8) & (path, 8) \\
$t_{13}$ & (path, 10) & (path, 10) \\
$t_{14}$ & (path, 12) & (path, 12) \\
$t_{15}$ & (path, 14) & (path, 14) \\
$t_{16}$ & (path, 16) & (path, 16) \\
$t_{17}$ & (peel, $\varnothing$) & (peel, $\varnothing$) \\
$t_{18}$ & (capture, $\varnothing$) & (capture, $\varnothing$) \\
$t_{19}$ & (end, $\varnothing$) & (end, $\varnothing$) \\
\hline
\end{tabular}
\end{table}

\textbf{Learning Action Effectiveness:} Our experiments reveal a significant decline in the efficacy of path actions as plans progress. Specifically, the improvements in average height diminish from more than 80\% reduction after the initial path to a below 30\% reduction after the eighth path. This suggests that the accuracy of paths becomes increasingly crucial in later stages of draping. Subsequent to the execution of initial paths, the uncompacted regions tend to concentrate near or on the edges. Analysis of $G_{2i}$ indicates that for both sheets, the orientation of uncompacted regions tends to be orthogonal to the edge as more and more compaction paths are executed. This can be observed in Fig. \ref{fig:sheet_status}. 
\\
\textbf{Selecting Search Parameters:}
From our search algorithm we observe that the top three actions with the best immediate state in the search tree generally lead to the best overall resulting state. Furthermore, increasing the value for parameter \(d_f\) beyond 4 causes the Gaussians to deviate significantly from observations, indicating an increase in process noise. Additionally, the system demonstrates a preference for exploring actions in $\mathbb{A}$ that are similar to the expert-crafted initial plans. For generating the refined plans, we set \(d_f=3\) and \(b_f=4\).
\\
\textbf{Refined Plans:} We get the refined plans for each sheet as shown in Table \ref{tab:refined-plans}. Fig. \ref{fig:paths_used} illustrates the paths in the final refined plans. We use the refined plans and conduct three experiments on the corresponding sheet. The experimental results are summarized in Table \ref{tab:results}. In comparison to the best expert-crafted plans, our system demonstrated about 40\% reduction on average in the total number of paths required for both sheets. Table \ref{table:comparison} provides a summary of the comparison between the two.

\begin{table}[!h]
\centering
\caption{Refined Plans for $\mathbf{sheet_1}$ and $\mathbf{sheet_2}$}
\label{tab:refined-plans}
\begin{tabular}{|c|c|c|}
\hline
\textbf{Time Step (t)} & \textbf{$\mathbf{\hat{D}}$ for $\mathbf{sheet_1}$} & \textbf{$\mathbf{\hat{D}}$ for $\mathbf{sheet_2}$} \\
\hline
$t_{1}$ & (path, 3) & (path, 7) \\
$t_{2}$ & (path, 11) & (path, 15) \\
$t_{3}$ & (path, 7) & (path, 5) \\
$t_{4}$ & (path, 15) & (path, 1) \\
$t_{5}$ & (path, 1) & (path, 13) \\
$t_{6}$ & (path, 9) & (path, 9) \\
$t_{7}$ & (peel, $\varnothing$) & (peel, $\varnothing$) \\
$t_{8}$ & (path, 5) & (path, 3) \\
$t_{19}$ & (path, 13) & (path, 11) \\
$t_{10}$ & (refine, 6) & (refine, 4) \\
$t_{11}$ & (capture, $\varnothing$) & (capture, $\varnothing$) \\
$t_{12}$ & (end, $\varnothing$) & (end, $\varnothing$) \\
\hline
\end{tabular}
\end{table}

\begin{table}[htbp]
\centering
\caption{Comparison between expert plan and optimal plan}
\label{table:comparison}
\begin{tabular}{|C{0.8cm}|C{2cm}|C{2cm}|C{1.6cm}|}
\hline
\textbf{Sheet} & \textbf{Average Paths in the Best Initial Plan} & \textbf{Average Paths in Refined Plans} & \textbf{Improvement} \\ \hline
\(sheet_1\) & 34.3 & 20.0 & 41.7\% \\ \hline
\(sheet_2\) & 27.3 & 16.3 & 40.3\% \\ \hline
\end{tabular}
\end{table}

\addtolength{\textheight}{-0.0 cm}   

\section{CONCLUSIONS}

\noindent
Our proposed framework presents a solution to the challenges inherent in generating draping plans for composite sheet layup. The primary focus of our approach is to refine initial layup paths with the specific goal of maximizing the time efficiency of the process while minimizing the leftover uncompacted regions. This refinement process integrates both human expertise and data driven decision making and learning, for a comprehensive solution that enhances adaptability — a critical requirement for the varying demands across production environments. Results show that our framework proves its effectiveness by reducing dependency on the correction cycles. This reduction leads to an overall enhancement in the time efficiency of robotic layup of composite sheets. The presented approach outperforms purely expert-crafted plans.

Although our framework demonstrates promising results, it is important to acknowledge several limitations. Firstly, the number of experiments performed place a limitation on the achievable accuracy and improvements. Secondly, our reliance on historical data for learning introduces the risk of biases or inaccuracies, necessitating continuous model updating and refinement. In future iterations, we aim to address these limitations by incorporating more comprehensive techniques. Specifically, we plan to integrate full-blown Gaussian Mixture Models (GMMs) to accurately model the entire mold. However, this approach would require a significant increase in the number of experiments conducted. Moreover, we intend to explore learning from human demonstrations to better model human behavior, rather than solely relying on sequencing from a limited set of paths. This broader approach could further enhance the adaptability and generalizability of our framework in real-world production environments.
\\ \\
{\bf Acknowledgments:} This research was sponsored by the ARM (Advanced Robotics for Manufacturing) Institute through a grant from the Office of the Secretary of Defense and was accomplished under Agreement Number W911NF-17-3-0004. The views and conclusions contained in this document are those of the authors and should not be interpreted as representing the official policies, either expressed or implied, of the Office of the Secretary of Defense or the U.S. Government. The U.S. Government is authorized to reproduce and distribute reprints for Government purposes notwithstanding any copyright notation herein.
\\ \\
We thank Brian Smith, Erik Wienhold, Christopher Brown, Ricardo Fritzke, and Sandra Zmeu from Boeing; Marc Simpson and Stan Fast from 3M Advanced Materials Division; and Miguel Rodriguez from ARM Institute for their support. We also acknowledge the use of Generative AI tools, Grammarly and ChatGPT, for light editing to correct grammatical errors.

\bibliographystyle{IEEEtran}
\bibliography{references}

\begin{thebibliography}{10}
\providecommand{\url}[1]{#1}
\csname url@samestyle\endcsname
\providecommand{\newblock}{\relax}
\providecommand{\bibinfo}[2]{#2}
\providecommand{\BIBentrySTDinterwordspacing}{\spaceskip=0pt\relax}
\providecommand{\BIBentryALTinterwordstretchfactor}{4}
\providecommand{\BIBentryALTinterwordspacing}{\spaceskip=\fontdimen2\font plus
\BIBentryALTinterwordstretchfactor\fontdimen3\font minus \fontdimen4\font\relax}
\providecommand{\BIBforeignlanguage}[2]{{%
\expandafter\ifx\csname l@#1\endcsname\relax
\typeout{** WARNING: IEEEtran.bst: No hyphenation pattern has been}%
\typeout{** loaded for the language `#1'. Using the pattern for}%
\typeout{** the default language instead.}%
\else
\language=\csname l@#1\endcsname
\fi
#2}}
\providecommand{\BIBdecl}{\relax}
\BIBdecl

\bibitem{compositesmanufacturing2016}
\BIBentryALTinterwordspacing
``State of the composites industry,'' Composites Manufacturing, 2016, accessed: [Insert access date here]. [Online]. Available: \url{http://compositesmanufacturingmagazine.com/2016/01/state-of-the-composites-industry-lucintel-2016}
\BIBentrySTDinterwordspacing

\bibitem{elkington2015hand}
M.~Elkington, D.~Bloom, C.~Ward, A.~Chatzimichali, and K.~Potter, ``Hand layup: {Understanding} the manual process,'' \emph{Advanced Manufacturing: Polymer \& Composites Science}, vol.~1, no.~3, pp. 138--151, 2015.

\bibitem{malhan2019determining}
R.~K. Malhan, A.~M. Kabir, B.~Shah, T.~Centea, and S.~K. Gupta, ``Determining feasible robot placements in robotic cells for composite prepreg sheet layup,'' in \emph{Proc. {ASME} Int. Manufacturing Science and Engineering Conf.}, vol.~1.\hskip 1em plus 0.5em minus 0.4em\relax American Society of Mechanical Engineers, 2019, pp. V001T02A025--1--V001T02A025--10.

\bibitem{frketic2017automated}
J.~Frketic, T.~Dickens, and S.~Ramakrishnan, ``Automated manufacturing and processing of fiber-reinforced polymer (frp) composites: An additive review of contemporary and modern techniques for advanced materials manufacturing,'' \emph{Additive Manufacturing}, vol.~14, pp. 69--86, 2017.

\bibitem{schuster2017autonomous}
A.~Schuster, M.~Kupke, and L.~Larsen, ``Autonomous manufacturing of composite parts by a multi-robot system,'' \emph{Procedia Manufacturing}, vol.~11, pp. 249--255, 2017.

\bibitem{manyar2022visual}
O.~M. Manyar, A.~Kanyuck, B.~Deshkulkarni, and S.~K. Gupta, ``Visual servo based trajectory planning for fast and accurate sheet pick and place operations,'' in \emph{International Manufacturing Science and Engineering Conference}, vol. 85802.\hskip 1em plus 0.5em minus 0.4em\relax American Society of Mechanical Engineers, 2022, p. V001T04A019.

\bibitem{malhan2021automated}
R.~K. Malhan, A.~V. Shembekar, A.~M. Kabir, P.~M. Bhatt, B.~Shah, S.~Zanio, S.~Nutt, and S.~K. Gupta, ``Automated planning for robotic layup of composite prepreg,'' \emph{Robotics and computer-integrated manufacturing}, vol.~67, p. 102020, 2021.

\bibitem{bjornsson2018automated}
A.~Bj{\"o}rnsson, M.~Jonsson, and K.~Johansen, ``Automated material handling in composite manufacturing using pick-and-place systems--a review,'' \emph{Robotics and Computer-Integrated Manufacturing}, vol.~51, pp. 222--229, 2018.

\bibitem{elkington2017automated}
M.~Elkington, C.~Ward, and A.~Sarkytbayev, ``Automated composite draping: a review,'' in \emph{SAMPE SEATTLE 2017}.\hskip 1em plus 0.5em minus 0.4em\relax SAMPE North America, 2017.

\bibitem{molfino2014design}
R.~Molfino, M.~Zoppi, F.~Cepolina, J.~Yousef, and E.~Cepolina, ``Design of a hyper-flexible cell for handling 3d carbon fiber fabric,'' \emph{Recent advances in mechanical engineering and mechanics}, vol. 165, 2014.

\bibitem{gambardella2022defects}
A.~Gambardella, V.~Esperto, F.~Tucci, and P.~Carlone, ``Defects reduction in the robotic layup process,'' \emph{Key Engineering Materials}, vol. 926, pp. 1437--1444, 2022.

\bibitem{mcconachie2020manipulating}
D.~McConachie, A.~Dobson, M.~Ruan, and D.~Berenson, ``Manipulating deformable objects by interleaving prediction, planning, and control,'' \emph{The International Journal of Robotics Research}, vol.~39, no.~8, pp. 957--982, 2020.

\bibitem{manyar2021simulation}
O.~M. Manyar, J.~Desai, N.~Deogaonkar, R.~J. Joesph, R.~Malhan, Z.~McNulty, B.~Wang, J.~Barbi{\v{c}}, and S.~K. Gupta, ``A simulation-based grasp planner for enabling robotic grasping during composite sheet layup,'' in \emph{2021 IEEE International Conference on Robotics and Automation (ICRA)}.\hskip 1em plus 0.5em minus 0.4em\relax IEEE, 2021, pp. 930--937.

\bibitem{ehinger2014robot}
C.~Ehinger and G.~Reinhart, ``Robot-based automation system for the flexible preforming of single-layer cut-outs in composite industry,'' \emph{Production Engineering}, vol.~8, no.~5, pp. 559--565, 2014.

\bibitem{zhang2018optimizing}
P.~Zhang, Z.~Zhou, G.~Chen, and S.~Chen, ``Optimizing the lay-up of composite tapes based on improved geodesic strategy for automated tape placement,'' \emph{Proceedings of the Institution of Mechanical Engineers, Part C: Journal of Mechanical Engineering Science}, vol. 232, no.~22, pp. 4084--4097, 2018.

\bibitem{yan2014accurate}
L.~Yan, Z.~C. Chen, Y.~Shi, and R.~Mo, ``An accurate approach to roller path generation for robotic fibre placement of free-form surface composites,'' \emph{Robotics and Computer-Integrated Manufacturing}, vol.~30, no.~3, pp. 277--286, 2014.

\bibitem{wang2023general}
K.~Wang, X.~Wang, J.~Gan, and S.~Jiang, ``A general method of trajectory generation based on point-cloud structures in automatic fibre placement,'' \emph{Composite Structures}, vol. 314, p. 116976, 2023.

\bibitem{gao2018optimal}
J.~Gao, ``Optimal motion planning in redundant robotic systems for automated composite lay-up process,'' Ph.D. dissertation, {\'E}cole centrale de Nantes, 2018.

\bibitem{chen2023multisensor}
L.~Chen, X.~Yao, K.~Liu, C.~Tan, and S.~K. Moon, ``Multisensor fusion-based digital twin in additive manufacturing for in-situ quality monitoring and defect correction,'' \emph{Proceedings of the Design Society}, vol.~3, pp. 2755--2764, 2023.

\bibitem{elkington2021real}
M.~Elkington, E.~Almas, B.~Ward-Cherrier, N.~Pestell, J.~Lloyd, C.~Ward, and N.~Lepora, ``Real time defect detection during composite layup via tactile shape sensing,'' \emph{Science and Engineering of Composite Materials}, vol.~28, no.~1, pp. 1--10, 2021.

\bibitem{manyar2022synthetic}
O.~M. Manyar, J.~Cheng, R.~Levine, V.~Krishnan, J.~Barbi{\v{c}}, and S.~K. Gupta, ``Synthetic image assisted deep learning framework for detecting defects during composite sheet layup,'' in \emph{International Design Engineering Technical Conferences and Computers and Information in Engineering Conference}, vol. 86212.\hskip 1em plus 0.5em minus 0.4em\relax American Society of Mechanical Engineers, 2022, p. V002T02A005.

\bibitem{tang2022process}
Y.~Tang, Q.~Wang, L.~Cheng, J.~Li, and Y.~Ke, ``An in-process inspection method integrating deep learning and classical algorithm for automated fiber placement,'' \emph{Composite Structures}, vol. 300, p. 116051, 2022.

\bibitem{chen2021rapid}
L.~Chen, X.~Yao, P.~Xu, S.~K. Moon, and G.~Bi, ``Rapid surface defect identification for additive manufacturing with in-situ point cloud processing and machine learning,'' \emph{Virtual and Physical Prototyping}, vol.~16, no.~1, pp. 50--67, 2021.

\bibitem{zhu2022challenges}
J.~Zhu, A.~Cherubini, C.~Dune, D.~Navarro-Alarcon, F.~Alambeigi, D.~Berenson, F.~Ficuciello, K.~Harada, J.~Kober, X.~Li \emph{et~al.}, ``Challenges and outlook in robotic manipulation of deformable objects,'' \emph{IEEE Robotics \& Automation Magazine}, vol.~29, no.~3, pp. 67--77, 2022.

\bibitem{yan2021learning}
W.~Yan, A.~Vangipuram, P.~Abbeel, and L.~Pinto, ``Learning predictive representations for deformable objects using contrastive estimation,'' in \emph{Conference on Robot Learning}.\hskip 1em plus 0.5em minus 0.4em\relax PMLR, 2021, pp. 564--574.

\bibitem{mcconachie2020bandit}
D.~McConachie and D.~Berenson, ``Bandit-based model selection for deformable object manipulation,'' in \emph{Algorithmic Foundations of Robotics XII: Proceedings of the Twelfth Workshop on the Algorithmic Foundations of Robotics}.\hskip 1em plus 0.5em minus 0.4em\relax Springer, 2020, pp. 704--719.

\bibitem{hu2018three}
Z.~Hu, P.~Sun, and J.~Pan, ``Three-dimensional deformable object manipulation using fast online gaussian process regression,'' \emph{IEEE Robotics and Automation Letters}, vol.~3, no.~2, pp. 979--986, 2018.

\bibitem{caccamo2016active}
S.~Caccamo, P.~G{\"u}ler, H.~Kjellstr{\"o}m, and D.~Kragic, ``Active perception and modeling of deformable surfaces using gaussian processes and position-based dynamics,'' in \emph{2016 IEEE-RAS 16th International Conference on Humanoid Robots (Humanoids)}.\hskip 1em plus 0.5em minus 0.4em\relax IEEE, 2016, pp. 530--537.

\bibitem{thach2022learning}
B.~Thach, B.~Y. Cho, A.~Kuntz, and T.~Hermans, ``Learning visual shape control of novel 3d deformable objects from partial-view point clouds,'' in \emph{2022 International Conference on Robotics and Automation (ICRA)}.\hskip 1em plus 0.5em minus 0.4em\relax IEEE, 2022, pp. 8274--8281.

\bibitem{jia2018learning}
B.~Jia, Z.~Hu, Z.~Pan, D.~Manocha, and J.~Pan, ``Learning-based feedback controller for deformable object manipulation,'' \emph{arXiv preprint arXiv:1806.09618}, 2018.

\bibitem{liu2023deformer}
D.~Liu, X.~Yu, M.~Ye, Q.~Zhangli, Z.~Li, Z.~Zhang, and D.~N. Metaxas, ``Deformer: Integrating transformers with deformable models for 3d shape abstraction from a single image,'' in \emph{Proceedings of the IEEE/CVF International Conference on Computer Vision}, 2023, pp. 14\,236--14\,246.

\end{thebibliography}

\end{document}